%% file: paper.tex
\documentclass[letterpaper, 10 pt, conference]{ieeeconf}  % Comment this line out if you need a4paper

\IEEEoverridecommandlockouts                              % This command is only needed if 
\title{\LARGE \bf
Offline Skill Graph (OSG): A Framework for Learning and Planning using Offline Reinforcement Learning Skills
}

\author{Ben-ya Halevy$^{1}$, Yehudit Aperstein$^{1}$, Dotan Di Castro$^{2}$% <-this % stops a space
\thanks{$^{1}$Ben-ya Halevy and Yehudit Aperstein are with  Intelligent Systems Department, Afeka College of Engineering, Tel-Aviv, Israel  {\tt\small benyai@mail.afeka.ac.il}}%
\thanks{$^{2}$Dotan Di~Castro is with Bosch Center for AI, Haifa, Israel
        {\tt\small dotan.dicastro@il.bosch.com}}%
}

\usepackage{float}
\usepackage{xcolor}
\usepackage{graphicx}
\usepackage{subfigure}
\usepackage{wrapfig}
\usepackage{algorithm}
\usepackage{algpseudocode}
\usepackage{amsmath}

\begin{document}

\maketitle
\thispagestyle{empty}
\pagestyle{empty}

%%%%%%%%%%%%%%%%%%%%%%%%%%%%%%%%%%%%%%%%%%%%%%%%%%%%%%%%%%%%%%%%%%%%%%%%%%%%%%%%
\input{abstract}
\input{introduction}
\input{related_work}
\input{setup}
\input{method}
\input{experiments}
\input{summary_future_work}

\bibliographystyle{IEEEtran}
\bibliography{paper}

\end{document}

%% file: abstract.tex
\begin{abstract}
Reinforcement Learning has received wide interest due to its success in competitive games. Yet, its adoption in everyday applications is limited (e.g. industrial, home, healthcare, etc.). In this paper, we address this limitation by presenting a framework for planning over offline skills and solving complex tasks in real-world environments. Our framework is comprised of three modules that together enable the agent to learn from previously collected data and generalize over it to solve long-horizon tasks. We demonstrate our approach by testing it on a robotic arm that is required to solve complex tasks.
\end{abstract}

%% file: introduction.tex
\section{INTRODUCTION}
\label{sec:introduction}

Reinforcement learning (RL; \cite{sutton2018reinforcement}) has shown great promise in the last decade, gaining popularity by surpassing professionals in complex games like “GO” \cite{silver2017mastering} and competitive video games \cite{mnih2015human, bellemare2013arcade}. These impressive capabilities have sparked the vision of robots learning and solving problems in the real world without the need for the coding of an implicit solution for each task.
However, although some real-world applications have been shown, the use of reinforcement learning in real-world robotic setups is not as wide spread as one would expect.

This is the result of three major issues. The first is the need for a long period of environment exploration, especially in continuous environments, where the state space is theoretically infinite. While this requirement is relatively easy to fulfill in games and simulations, exploration in the real world takes a great amount of time and there is a high probability of damage to the robot, environment, or both. The second issue is that most RL algorithms are designed to solve one task at a time and need to restart the learning process from scratch for any new problem, even when it is defined on the very same environment. The last issue is the complexity of the solutions and low level of explainability when deep neural networks are applied to complex tasks. The ability to explain the agent's decisions is important for understanding what drives it to act in a certain way in different scenarios, and for improving specific aspects directly instead of going in blindly. Also, explainability is important for safety measures in real-world scenarios and is often a prerequisite for allowing a robot to use an AI-based solution.

Several recent works have tackled these issues, usually focusing on only one of them without addressing the others. One proposed approach is offline RL (ORL; \cite{levine2020offline}), which avoids extensive exploration by allowing the agent to learn a policy from a static data set. Yet, ORL is still mostly limited to a single task per policy. By connecting learned skills in a graph, the deep skill graphs approach (DSG; \cite{bagaria2021skill}) enables the use of planning over different tasks. This multi-goal solution is more explainable than RL alone, but requires great exploration and does not address decision making in states that are not on the graph. 

\begin{figure}[H]
\includegraphics[scale = 0.29]{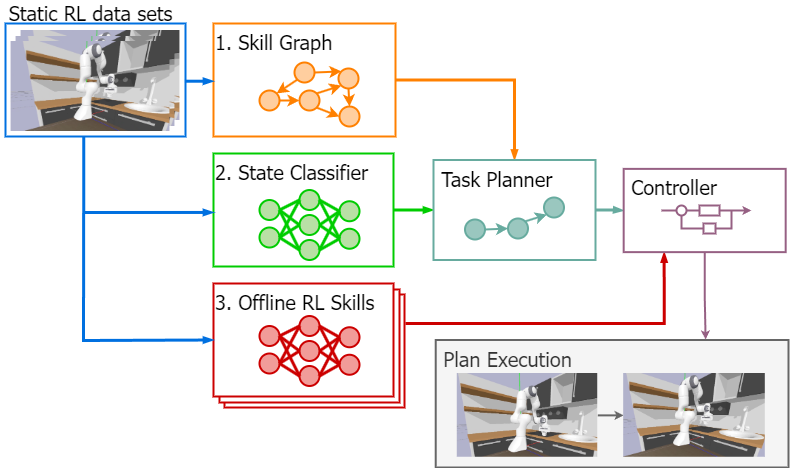}
\centering
\caption{Overview of the proposed framework. The framework consists of three major modules: a set of ORL skills, a deep skill graph, and a state-to-skill classifier. Combined with a planner and a controller, this setup provides the means to solve complex new tasks without training on the entire trajectory beforehand.}
\label{fig:Overview}
\end{figure}

We believe that in order to utilize a data-driven solution for real-world robotic tasks, the solution needs to address all three above-noted issues simultaneously. In this paper, we  suggest a framework for using ORL skills in complex real-world robotic scenarios, in combination with a planner and a controller (see a schematic overview in Fig. \ref{fig:Overview}). This framework is comprised of three major modules: a set of offline learned skills, an offline deep skill graph, built on top of these skills, and a state classification Network, which connects our state space to our graph through reliable first skill selection. These components are learned in an offline manner using a previously collected data set, and can then be used to effectively plan and execute solutions for different tasks and grow over the course of lifetime of the agent. 

Our main contributions in this work are as follows:
\begin{enumerate}
    \item A framework for using ORL with DSG for robotics.
    \item A method for state classification and planning over skills.
    \item A demonstration of our solution over complex motion-planning tasks.
\end{enumerate}

The paper is organized as follows. In Section \ref{sec:related_work}, 
we review related work, while in Section \ref{sec:setup}, we describe the problem's setup. In Section \ref{sec:method}, we detail the methodology, and In Section \ref{sec:experiments}, we show the applicability of our work in simulation. We conclude in Section \ref{sec:summary_future_work}, pointing out future work directions.

%% file: related_work.tex
\section{RELATED WORK}\label{sec:related_work}
A number of approaches have been proposed  for overcoming the hurdles that prevent the wide implementation of reinforcement learning in real-world robotic setups. In this section, we  review the most relevant and recent ones while highlighting the differences between them and the work presented in this paper.

\textbf{Macro Actions and Options:} The first approach that was proposed for addressing the exploration and learning times in complex RL environments is the implementation of macro actions. A macro-action is a sequence of actions chosen from the primitive actions of the problem. Lumping actions together as macros could greatly aid in solving large problems \cite{korf1983learning, gullapalli1992reinforcement} and sometimes greatly expedites learning \cite{iba1989heuristic, mcgovern1997roles, mcgovern1998macro, sutton1998between, sutton1999improved}.

A more recent approach is the options framework \cite{sutton1999between}, which involves the generalization of primitive actions to include temporally extended courses of action. This framework consists of three components: a policy $\pi:\ S\times\mathrm A\ \rightarrow\ [0,1]$, a termination condition $\beta:\ S\ \mathrm\rightarrow\ [0,1]$, and an input set $I\ \mathrm{\subseteq\ S}$. An option is available in state s if and only if $s\in I$. If the option is taken, then actions are selected according to $\pi$ until the option terminates stochastically according to $\beta$. Note that primitive actions can be viewed as a special case of options.

\textbf{Skill Chains:} The growing popularity of the options framework led to the development of several other methods that leverage the framework even further and accelerate learning in continuous spaces. One such approach is Skill Chaining \cite{konidaris2009skill}, which tries to decompose a continuous problem into a chain of skills and then uses the option framework to learn those skills. The chaining procedure is done by defining each previous skill’s initial states as goal states of the current skill. This approach defines a goal skill, which indicates how the agent can reach the goal state once it is in close proximity to the goal region. Then, through the skill discovery process, the initiation set of the current skill is considered as the target set of another skill and a new option is defined to reach this target set. This procedure is repeated until the start state is reached, at which point a chain of skills has been constructed.

In \cite{konidaris2011cst}, the authors introduce a method to segment expert demonstrations into skills and then chain these skills into trees, which branch out toward the goal through different trajectories. We here propose to expand this idea to graphs. The representation of a tree opens the possibility of using search-based planning, and adopting a graph representation means that our agent can start at any given state and reach any given goal represented by the skill graph. 

\textbf{Planning over Skills:} Skills (options) have also been shown to empirically speed up planning in several domains \cite{silver2012compositional, jinnai2019finding, james2018learning, francis1993utility, konidaris2016constructing}. Recently, \cite{sharma2019dynamics} introduced the DADS algorithm, which uses model-free RL methods to learn stochastic models for specific behaviors (skills). After learning several behaviors, model predictive control is used to plan the latent space of those behaviors. This approach has been shown to outperform both the standard model-based RL and model-free goal-conditioned RL. Yet, while promising in the context of complex movement, this skill-learning method seems computationally heavy and the result likewise requires the use of a computationally heavy planning method. Moreover, the experiments conducted involved only movement in an empty environment with no obstacles or interactions with other objects. We aim to represent skills in a much simpler fashion, allowing for a much more straightforward planning scheme, using graph-based planning. Our scheme will enable not only zero-shot planning but also to take into account different environments and different types of tasks.

Another example of planning over skills is Deep Skill Graphs. This scheme was proposed in \cite{bagaria2021skill}, where the authors used exploration to discover skills and connect them in a skill graph as a means to plan the movement of an agent inside a maze. This method was not tested in a more realistic environment and, to date, does not include interactions with movable objects. We aim to employ the same approach, but to make it suited to more real-life problems where agents cannot always freely explore their environment without causing damage to it or themselves. In our approach, data can be either collected in a controlled fashion or received from past experiments, even those made by others, and the skills can be tested or executed on the environment after the learning process. 

Graphs can also be useful for life-long solutions for robotic agents. In \cite{liang2021search}, the authors demonstrated a method for building a graph that involves simulating pre-gained skills and learning a skill effect model. This model can predict the cost and effect of each skill and improve the agent's planning capabilities while adding skills over time. With our approach, skill effects are derived from the offline data sets that are used, meaning that they do not need to be simulated to learn the effect after training. This also applies to calculating cost, an issue we do not address in this research.

\textbf{Offline Reinforcement Learning:} A number of recently proposed ORL algorithms that eliminate the need for exploration all together have shown promise both in simulation and in real-life applications \cite{levine2020offline}. Some of them have even produced state-of-the-art (SOTA) results by either regularizing the policy \cite{fujimoto2021minimalist} or the Q function \cite{kumar2020conservative} in known online Actor-Critic models. In \cite{singh2020cog}, the authors showed that combining diverse data sets with data related to a specific skill can improve offline learning and allow skill chaining, with only a sparse binary reward for task completion. In another work that shares our motivation  \cite{lynch2019play}, the authors suggest a hierarchical learning method that consists of a high-level goal-setting policy and a low-level subgoal-conditioned policy, for solving several long horizon tasks using imitation learning. The suggested solution is able to tackle a number of long horizon tasks in simulation, after an extensive online fine tuning process, but is unable to reach sufficient success rates purely from imitation. Our work shows that better offline performance can be achieved by utilizing ORL methods for learning different low level skills and planning over them, for solving long horizon tasks. Our solution for using a planner instead of a goal setting policy is more reliable, scalable and does not require learning, allowing the framework to grow and develop over the life cycle of the agent.

%% file: setup.tex
\section{Setup}
\label{sec:setup}
In this section, we describe the setup we use throughout the paper. Consider a Reinforcement Learning agent \cite{sutton2018reinforcement} that interacts with an environment over time. At each time step, $t$, the agent perceives a state, $s_t$, in state space $S$ and selects an action, $a_t$, from action space $A$, by following policy $\pi(a_t|s_t)$, which is the agent’s behavior, i.e., a mapping from state $s_t$ to actions $a_t$. The agent then receives a scalar reward, $r_t$, determined by the reward function, $R(s,a)$, and transitions to the next state, $s_{t+1}$, according to the environment's dynamics, or model, determined by the state transition probability, $P{(s}_{t+1}|s_{t,}a_t)$. 

In an episodic problem, this process continues until the agent reaches a terminal state, and then it restarts. We define the return as the discounted accumulated rewards, $R_t=\ \sum_{k=0}^{\infty}{\gamma^tr_{t+k}}$, where the discount factor $\gamma\in\left(0,1\right]$. The agent's goal is to maximize the expectation of the long-term return. We note that this formulation fits the discrete state's and action spaces' representation, and it can also be generalized to continuous spaces.

The Offline Reinforcement Learning (\cite{levine2020offline}) problem can be defined as a data-driven formulation of the RL problem. While the end goal is still to maximize the return, in ORL the agent no longer has the ability to interact with the environment and collect additional transitions using the behavior policy. Instead, the learning algorithm is provided with an offline data set of transitions, $D = {(s^i_t, a^i_t, s^i_{t+1}, r^i_t)}$, and the agent is required to optimize the policy using this data set. This formulation resembles the standard supervised learning problem statement, and we can regard $D$ as the training set for the policy. In essence, ORL requires the learning algorithm to reach a sufficient understanding of the dynamical system solely from a fixed data set, and then to construct a policy that attains the largest possible cumulative reward when it is actually used to interact with it. 

%% file: method.tex
\section{METHOD} \label{sec:method}
In this section, we describe our method for implementing the three modules of our framework (described in Fig. \ref{fig:Overview}). 

\textbf{ORL Skills Module.} For the  offline learning process, a data set needs to be labeled with rewards according to the goal states of each trajectory. For example, for picking up an object a robot must first reach the objects area, grab the object and then lift it up. Each of these parts can be considered as a trajectory of a skill.  We begin with a static (offline) data set and break it into multiple data sets of connected trajectories. For an existing large data set, this can be done manually by selecting useful intersections, automatically by using algorithms such as Relay Data Relabeling \cite{lynch2019play} or by using unsupervised learning methods to create different data clusters. Another method is to build a data set by collecting smaller data sets for specific skills that are relevant for the agent, using simple noisy policies. We demonstrate this method in the Experiments Section (Section \ref{sec:experiments}). 

Once a set of labeled samples is generated, the agent is trained to learn the skills using state-of-the-art ORL algorithms such as TD3+BC \cite{fujimoto2021minimalist}, CQL \cite{kumar2020conservative} or others. Then, we use the options framework \cite{sutton1999between} to represent our policies and chain them together. Each option consists of the tuple $<I,\pi,\beta>$, where $I$ is the initial state where the option is available, $\pi$ is the policy that was learned using the ORL, and $\beta$ is the termination state, i.e., the goal state the policy aims to reach. 

\textbf{Skills Graph Module.} The SGM is built based on the ORL skills module, by chaining options \cite{konidaris2009skill, konidaris2011cst}, such that one termination state, $\beta$, is another option's initial state, $I$, if such a state exists. Additionally, any effect on the environment that is not within the agent's state space is added to the relevant node. For instance, if a door is opened according to the policy and the state space does not include that door, then the next node will note that change in order to use it in the planning process. 

Our skill graph allows the agent to move around the environment between its state nodes. But what happens when the agent finds itself in a state that is nowhere on the graph? Since we are looking at a continuous state space, we know that there is a continuum  of states between the nodes of the graph. We note that learning more skills and adding more nodes will result in better coverage of the environment, but at the same time, it will increase the size of the graph with no real benefit to the ability to solve tasks.

\textbf{State Classifier Module.} In addition to utilizing an offline data set to learn skills, we propose to use it to train a simple classifier network as well. This is done by relabeling the data by skill, where each point in the state space $S$ is labeled with the skill class that used this point during training. Then, we simply run a fully connected classification network that takes a world state as input and returns the skill class most relevant to that (unseen) state. 

This classifier's purpose is to select the skill policy the agent utilizes to reach any node on the graph with high reliability. Note that it does not matter which node is selected, as long as the chance of reaching it is high. When The agent receives a task, the algorithm first checks whether the agent is in a state that is considered on the graph. If not, it classifies that state to find the best policy for reaching the graph, and then strive to reach the goal from that point. The planner plans a trajectory of skills in order to reach the goal state and also to introduce changes to the environment set by the task, if those are represented by the state space (e.g., a door's state set to "open").

%% file: experiments.tex
\section{Experiments} \label{sec:experiments}
\begin{figure}[H]
\includegraphics[scale = 0.15]{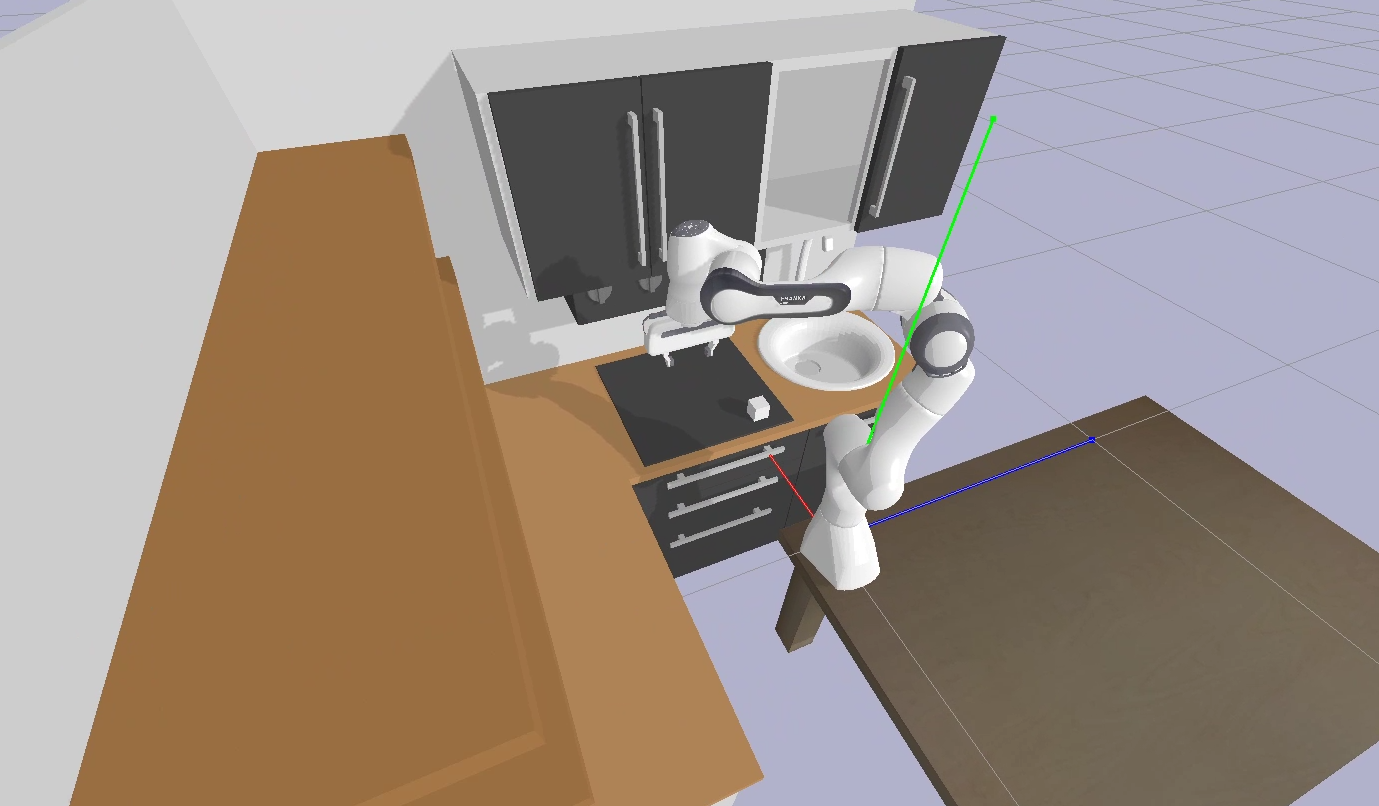}
\centering
\caption{Simulation environment of a corner kitchen with
a manipulator arm attached to a table, 2 shelves, cabinets
that can open and valves that can turn.}
\label{fig:Environment}
\end{figure}

We evaluate our framework on a simulated robot manipulator (robotic arm) environment built with PyBullet \cite{coumans2019}. This environment is comprised of a Franka Emika 7-degrees-of-freedom robot arm and an interactive kitchen with movable parts such as doors and valves. To enable free body interactions, we also position a small cube that can move freely on top of the kitchen's counter. The robot arm can move around and interact with, grab or manipulate objects inside the environment. Fig. \ref{fig:Environment} demonstrates the environment. 

\subsection{State and Action Spaces}
We define the agent's world state space as the states of our robot $$[[x_1, x_2, x_3], [q_1, q_2, q_3, q_4], [g1, g2, g3]],$$
where $[x_1,x_2,x_3]$ represent the robot's end effector position, $[q_1, q_2, q_3, q_4]$ the end effector's orientation quaternion, and $[g1, g2, g3]$ the state of the gripper: the given command, detected grip force and distance between the gripper's fingers, respectively.

With no other information from the environment, we separate the learned skills so as to only depend on the robot's current state and not be affected by changes to the environment. For simplicity, we do not use a mounted camera, though using a camera would help increase the policy's robustness in cases where the agent interacts with moving objects. Information about changes to the environment will be addressed in the graph for the planning process, given each skill's effect on it.

To control the robot, we use inverse kinematics (IK; \cite{goldenberg1985complete}), such that we only command the end effector's target position and orientation. The IK model then calculates each joint's target angle in order to reach the desired state, and these angles then serve as input for the joint angle controllers. In this control scheme, our agent's action space is of the following form: $$[\Delta x,\Delta y,\Delta z,\Delta\alpha,\Delta\beta,\Delta\gamma,gCmd]$$, where $[\Delta x,\Delta y,\Delta z]$ are the relative position commands, $[\Delta\alpha,\Delta\beta,\Delta\gamma]$ the relative orientation commands, and $gCmd$ a continuous value from $[-1, 1]$ that prompts the gripper to close
completely when less than -0.5 and to open completely when greater than 0.5. 

\subsection{Data Collection}
We used a scripted noisy policy that has a predefined success rate between 40\% and 50\%. This allowed us to quickly and easily acquire highly specific data sets, using a simple noisy control scheme, and gauge the improvement in the success rate over the noisy policies. This method was also shown to be useful in \cite{singh2020cog}.

Algorithm \ref{alg:Sequence} is a simple noisy policy that allows us to set any trajectory as a sequence of different states that are reached one after the other. The sequence of states can define any task, such as moving around, grabbing objects, moving objects and interacting with the environment.
The algorithm calculates the error in the position and the orientations of the robot, compared with the target state, and uses that error as a relative action to the robot control scheme.

In order to control the success rate of this policy, we change the threshold, number of time steps and amount of random noise added to the action at each step. At each time step in the loop, we inject the noisy action into the environment, which simulates a reaction to that action and returns the next observation and reward for the new world state. This step is repeated until the target number of recorded full episodes is reached. In our experiment, we use the spars binary reward format, in which a reward of 1 is recorded whenever the robot achieves the task given in our sequence. We show that binary rewards can be sufficient to learning policies using current SOTA algorithms.

\begin{algorithm}
\caption{Scripted sequence}\label{alg:Sequence}
\begin{algorithmic}[1]
\State Sequence $\gets$ $(state_1,..., state_n)$
\State threshold $\sim \mathcal{N}(\mu_1,\,\sigma_1^{2})$
\State numOfTimesteps $\gets$ number of time 
\For{state \textbf{in} Sequence}
    \State targetState $\gets$ state
    \State targetGripper $\gets$ state.gripper\_state
    \For{steps \textbf{in} (0,numOfTimesteps)}
            \State eeState $\gets$ end effector State
            \State stateDist $\gets$ distance(targetState, eeState)
            \State gripperState $\gets$ gripper state
            \If{stateDist $>$ threshold}
                \State action $\gets$ targetState $-$ eeState
            \ElsIf{gripperState $\neq$ targetGripper}
                \State action $\gets$ change gripper state
            \Else
                \State action $\gets$ 0
            \EndIf
            \State noise $\sim \mathcal{N}(\mu_2,\,\sigma_2^{2})$
            \State action $\gets$ action + noise
            \State $(s,r,s')$ $\gets$ env.step(action)
    \EndFor
\EndFor
\end{algorithmic}
\end{algorithm}

Using This method, we collect 12 data sets for our skills. Of these, 6 are for movement skills, where the robot changes position and orientation without interacting with the environment. Another 2 are for grabbing and sliding a door open, and the other 4 are for grabbing, picking up, moving and placing the cube located on the counter, in that order.

\subsection{Offline Skill Learning}
 After collecting the data sets, we used TD3+BC \cite{fujimoto2021minimalist} to learn all the above-noted skills. We chose this algorithm because it is very simple to use and requires shorter run times. Testing each policy after training highlights the improvement over the data collection policy. Even under very noisy conditions, our approach typically yields high success rates, as presented in Table \ref{table:skills}. While testing the skills, we apply a normally distributed noise $\mathcal{N}(0,\,\sigma^{2})$ to the observed position of the gripper, where $\sigma^{2}$ is equal to 1-2 cm (1-2\% of total range). These success rates are achieved using small-sized data sets of 10K simulation episodes at most and only several hours of training. The skills that are most sensitive are skills for precisely grabbing and placing the cube that is nearly the size of the gripper itself, which explains the degradation under high levels of noise.

\begin{table}[h!]
\caption{offline skills success rate tests}
\label{table:skills}
\begin{center}
 \begin{tabular}{|c||c|c|c|}
 \hline
  skill number & no noise & $\sigma^2$ = 1.0e-2 & $\sigma^2$ = 2.0e-2\\
 \hline
 \hline
  1 & 1.0 & 1.0 & 1.0 \\
 \hline
  2 & 1.0 & 1.0 & 1.0   \\
 \hline
  3 & 1.0 & 1.0   & 1.0 \\
 \hline
  4 & 1.0 & 1.0 & 1.0 \\
 \hline
  5 & 1.0 & 1.0 & 1.0 \\
 \hline
  6 & 1.0 & 0.99 & 0.99 \\
 \hline
  7 & 0.96 & 0.96 & 0.83  \\
 \hline
  8 & 1.0 & 0.94   & 0.28 \\
\hline
  9 & 1.0 & 1.0   & 1.0  \\
 \hline
  10 & 1.0 & 1.0 & 1.0  \\
 \hline
  11 & 0.99 & 0.98 & 0.98  \\
 \hline
  12 & 0.95 & 0.77 & 0.65  \\
 \hline
\end{tabular}
\end{center}
\end{table}

\subsection{Offline Deep Skill Graph}
\begin{figure}[h]
\includegraphics[scale = 0.37]{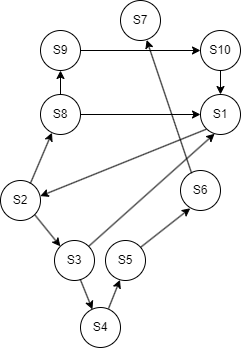}
\centering
\caption{Graphical representation of the resultant skill graph. %The nodes are placed in a way that somewhat resembles the geometry of the movement in the environment.
This graph includes skills for moving, opening a sliding cabinet door, picking up a cube, moving the cube and placing it inside an open cabinet.}
\label{fig:our_graph}
\end{figure}

As discussed in Section \ref{sec:method}, we created our graph based on the options framework, connecting the skills by their initial and termination states. In practice, we created a skills dictionary, where the keys are the initial states and the items were comprised of possible skills from those states. Each skill contained the termination state, the environment's effect, and the policy's index to the model's source. A graphical representation of our graph is shown in Fig. \ref{fig:our_graph}

When we run our planner, it starts forming the graph nodes from the starting state and uses the skill dictionary to create each node's child nodes, and so on, as it plans over them. The planner also keeps a record of the changes in the environment, searching for a combination of effects that was tasked by the user. 

\subsection{State-Skill Classifier}
After we finish creating our skills and skill graph, we turn to creating our state classifier, responsible for connecting our state space to our graph. As explained in Section \ref{sec:method}, with this classifier, we can start working with our robot without the need to manually start it in a state that exists on the graph. We teach our classifier to select the skill with the best chance of getting the robot to any node, which can then be used as a starting point for planning.

For practical reasons, we only use skills that have no direct effect on the environment, since all environmental effects should be a result of proper planning. Our experiment includes 6 movement skills, all of which should not interfere with our working environment. Thus, we create a classifier for six output classes.

Before we start training, we must first create a fitting data set. We extract from each of our previous data sets the state at each data point and attach the relevant skill tag as a label, according to the respective skill that was trained using each data set. To complete the process, we merge the 6 new data sets into one.

As for the network, we use a fully connected net with 2 hidden layers with 256 nodes in each. The inputs are the 10 state variables of our state space and the outputs are 6 values for each skill, where the chosen skill has the maximum value.

The training process is set to include 9 epochs; for each epoch, the classifier is trained on the training set in batches of 1,028 data points. At the end of the training, we test the model on the test set, which yields the confusion matrix in Table \ref{table:confusion_matrix}. The values in the table show the number of data points for each skill, classified to each of our classes and normalized to the total number of data points.

\begin{table}[h]
\caption{Classifier confusion matrix}
\label{table:confusion_matrix}
\begin{center}
\begin{tabular}{|c||c|c|c|c|c|c|}
 \hline
 skill & 1 & 2 & 3 & 4 & 5 & 6 \\
 \hline
 \hline
  1 & 0.129  & 0     & 0.002  & 0.007  & 0      & 0      \\
 \hline
  2 & 0      & 0.136 & 0      & 0      & 0      & 0.003  \\
 \hline
  3 & 0.002  & 0.005 & 0.12   & 0.001  & 0      & 0      \\
 \hline
  4 & 0      & 0      & 0     & 0.285  & 0.028  & 0.012  \\
 \hline
  5 & 0      & 0      & 0     & 0.034  & 0.103  & 0.002  \\
 \hline
  6 & 0      & 0      & 0     & 0.043  & 0.017  & 0.067  \\
 \hline
\end{tabular}
\end{center}
\end{table}

Normally, an accuracy score would be the sum of the main diagonal values of the confusion matrix. Applying this summation action to the resulting matrix yields an accuracy score of 0.84, or 84\%. Even though this is not a bad score, we have to remember the objective of the suggested classifier---which is solely to successfully direct the agent to any node on the skill graph. With this in mind, we arrange the classified skills so that skills that start or end in the same initial or termination state are next to each other. This allows us to use a corrected accuracy score, which also sums the values along the two diagonals adjacent to the main diagonal. Under this scheme, both skills are selected when the agent is around the state that intersects the said two skills. The resulting corrected accuracy score then rises to 0.927, or 92.7\%.

To prove the classifier's power, it is not enough to test its classification accuracy on the data. We must also run tests in simulation. We did so by starting the robot at random states in a radius of 15cm (15\% of full range) from 8 chosen world states, in orientations diverging by up to 0.2 radians in each angle. We then run the classifier and test the success rate of the robot in reaching the chosen skill's termination state. We do this for 1K random episodes around each state. The first 5 states are the states where our tested skills start or end. The last 3 states are located along the long edges of our graph, meaning that there is a far greater distance between the starting point and any other node state. As shown in Table \ref{table:classifier_results}, we obtain success rates of almost 100\%. This shows how powerful this classifier is at solving the problem of planning from random states in a continuous environment.

\begin{table}[h]
\caption{Classifier test results}
\label{table:classifier_results}
\begin{center}
\begin{tabular}{|c||c|}
 \hline
 Test number & Success rate \\
 \hline
 \hline
  1 & 0.96 \\
 \hline
  2 & 0.96 \\
 \hline
  3 & 1.0 \\
 \hline
  4 & 0.97 \\
 \hline
  5 & 1.0 \\
 \hline
  6 & 1.0 \\
 \hline
  7 & 1.0 \\
 \hline
  8 & 1.0 \\
 \hline
  Average & 0.99 \\
 \hline
\end{tabular}
\end{center}
\end{table}

\subsection{Full Pipeline Testing}
After implementing the entire framework, we test it on a few diverse tasks. We start with simple stand-alone tasks (First Experiment), like picking up the cube and opening a door separately. Then, we start combining the tasks until we run the entire process of opening the door, picking up the cube and placing the cube in the cabinet as one complete task (Second Experiment). For each test, we create a random start state, as we did in the previous subsection. At each starting state, we run the state classifier and receive our first skill and termination state. The skill is added as the first action of the plan and the planner receives the termination state as its starting state for planning and its goal state---the task's objective. The output plan is than passed to a low level controller, which executes it, skill by skill and decides when each skill ends (termination state achieved) and another begins. Since we are using an IK model, the skills can be badly effected by changes in the joint configuration. In order to alleviate these effects, the controller quickly resets the joint configuration after each skill is executed, so it is closer to the initial conditions for the next skill. The results of the two experiments are shown in Tables \ref{table:short_tasks} and \ref{table:combined}.

In the first experiment, we notice that even though the task remains the same, starting from different points in the world can effect the result, more so when the task is more complicated. This can be attributed to the use of IK to control the robot and to starting each test at a random state without checking every possibility for collision. Overall the tests yield high success rates and prove the framework is effective at solving multiple tasks, without using any additional online learning for fine tuning.

\begin{table}[h!]
\caption{first experiment - short tasks}
\label{table:short_tasks}
\begin{center}
 \begin{tabular}{|c||c|c|}
 \hline
  Test number & pick up & open door \\
 \hline
 \hline
  1 & 0.98 & 0.89\\
 \hline
  2 & 0.90 & 0.81\\
 \hline
  3 & 1.0  & 0.96\\
 \hline
  4 & 0.91 & 0.71\\
 \hline
  5 & 0.71 & 0.74\\
 \hline
  6 & 1.0  & 0.9\\
 \hline
  7 & 0.89 & 0.83\\
 \hline
  8 & 0.98 & 0.92\\
 \hline
  Average & 0.92 & 0.84 \\
 \hline
\end{tabular}
\end{center}
\end{table}

\begin{table}[h!]
\caption{second experiment - combined tasks}
\label{table:combined}
\begin{center}
 \begin{tabular}{|c||c|c|c|}
 \hline
  Test number & open+pickup & pickup+place & open+pickup+place\\
 \hline
 \hline
  1 & 0.89 & 0.87 & 0.83 \\
 \hline
  2 & 0.78 & 0.82 & 0.77 \\
 \hline
  3 & 0.95 & 0.92 & 0.87  \\
 \hline
  4 & 0.83 & 0.71 & 0.67 \\
 \hline
  5 & 0.73 & 0.75 & 0.67 \\
 \hline
  6 & 0.89 & 0.88 & 0.85 \\
 \hline
  7 & 0.87 & 0.77 & 0.77 \\
 \hline
  8 & 0.89 & 0.89 & 0.86  \\
 \hline
  Average & 0.85 & 0.82 & 0.78 \\
 \hline
\end{tabular}
\end{center}
\end{table}

%% file: summary_future_work.tex
\section{CONCLUSIONS} \label{sec:summary_future_work}
This work presents a scalable data-driven framework for using offline RL in life-long, real-world scenarios and demonstrates its potential on complex long-horizon tasks. We also introduce a new component, the state-to-skill classifier, and prove its value for planning over skills in a continuous environment. The suggested framework is able to perform well over different tasks, starting from random points in the environment and achieves better success rates overall, compared to previous works, based purely on offline learning.
Future work will focus on improving the framework's robustness and will also aim to miniaturize the framework and reduce the amount of policies required for movement, via a goal conditioned policy approach. Furthermore, we would like to tests the frameworks ability to reach states that are not on the graph by further utilizing the state-to-skill network and using offline and online learning to expand the skill graph over time.